\documentclass[conference]{IEEEtran}
\IEEEoverridecommandlockouts
\usepackage{cite}
\usepackage{amsmath,amssymb,amsfonts}
\usepackage{algorithmic}
\usepackage{graphicx}
\usepackage{textcomp}
\usepackage{xcolor}
\def\BibTeX{{\rm B\kern-.05em{\sc i\kern-.025em b}\kern-.08em
    T\kern-.1667em\lower.7ex\hbox{E}\kern-.125emX}}
\usepackage{soul}
\usepackage{subcaption}
\usepackage{lipsum}
\usepackage{scalerel}
\usepackage{tikz}
\usetikzlibrary{svg.path}

\definecolor{orcidlogocol}{HTML}{A6CE39}
\tikzset{
  orcidlogo/.pic={
    \fill[orcidlogocol] svg{M256,128c0,70.7-57.3,128-128,128C57.3,256,0,198.7,0,128C0,57.3,57.3,0,128,0C198.7,0,256,57.3,256,128z};
    \fill[white] svg{M86.3,186.2H70.9V79.1h15.4v48.4V186.2z}
                 svg{M108.9,79.1h41.6c39.6,0,57,28.3,57,53.6c0,27.5-21.5,53.6-56.8,53.6h-41.8V79.1z M124.3,172.4h24.5c34.9,0,42.9-26.5,42.9-39.7c0-21.5-13.7-39.7-43.7-39.7h-23.7V172.4z}
                 svg{M88.7,56.8c0,5.5-4.5,10.1-10.1,10.1c-5.6,0-10.1-4.6-10.1-10.1c0-5.6,4.5-10.1,10.1-10.1C84.2,46.7,88.7,51.3,88.7,56.8z};
  }
}

\newcommand\orcidiconKLW[1]{\href{https://orcid.org/0000-0002-1938-4222}{\mbox{\scalerel*{
\begin{tikzpicture}[yscale=-1,transform shape]
\pic{orcidlogo};
\end{tikzpicture}
}{|}}}}
\newcommand\orcidiconAAS[1]{\href{https://orcid.org/0000-0002-6140-619X}{\mbox{\scalerel*{
\begin{tikzpicture}[yscale=-1,transform shape]
\pic{orcidlogo};
\end{tikzpicture}
}{|}}}}
\newcommand\orcidiconAK[1]{\href{https://orcid.org/0000-0002-3494-0469}{\mbox{\scalerel*{
\begin{tikzpicture}[yscale=-1,transform shape]
\pic{orcidlogo};
\end{tikzpicture}
}{|}}}}
\newcommand\orcidiconFGS[1]{\href{https://orcid.org/0000-0002-5090-9007}{\mbox{\scalerel*{
\begin{tikzpicture}[yscale=-1,transform shape]
\pic{orcidlogo};
\end{tikzpicture}
}{|}}}}

\usepackage[bookmarks=false]{hyperref}

\begin{document}

\title{Feed-forward Disturbance Compensation for Station Keeping in Wave-dominated Environments\\
\thanks{This work was supported by the EPSRC under grant No. EP/R513209/1.}
}

\author{
\IEEEauthorblockN{Kyle L. Walker\IEEEauthorrefmark{1}\orcidiconKLW{0000-0002-1938-4222}, Adam A. Stokes\IEEEauthorrefmark{1}\orcidiconAAS{0000-0002-6140-619X}, Aristides Kiprakis\IEEEauthorrefmark{2}\orcidiconAK{0000-0002-3494-0469} and Francesco Giorgio-Serchi\IEEEauthorrefmark{1}\IEEEauthorrefmark{3}\orcidiconFGS{0000-0002-5090-9007}}
\IEEEauthorblockA{\IEEEauthorrefmark{1}Institute for Integrated Micro and Nano Systems}
\IEEEauthorblockA{\IEEEauthorrefmark{2}Institute for Energy Systems\\
School of Engineering\\
University of Edinburgh\\
Edinburgh, UK\\
\IEEEauthorrefmark{3}Correspondence: \textit{F.Giorgio-Serchi@ed.ac.uk}}}

\maketitle

\begin{abstract}
When deploying robots in shallow ocean waters, wave disturbances can be significant, highly dynamic and pose problems when operating near structures; this is a key limitation of current control strategies, restricting the range of conditions in which subsea vehicles can be deployed. To improve dynamic control and offer a higher level of robustness, this work proposes a Cascaded Proportional-Derivative (C-PD) with Feed-forward (FF) control scheme for disturbance mitigation, exploring the concept of explicitly using disturbance estimations to counteract state perturbations. Results demonstrate that the proposed controller is capable of higher performance in contrast to a standard C-PD controller, with an average reduction of $\approx 48\%$ witnessed across various sea states. Additional analysis also investigated performance when considering coarse estimations featuring inaccuracies; average improvements of $\approx 17\%$\% demonstrate the effectiveness of the proposed strategy to handle these uncertainties. The proposal in this work shows promise for improved control without a drastic increase in required computing power; if coupled with sufficient sensors, state estimation techniques and prediction algorithms, utilising feed-forward compensating control actions offers a potential solution to improve vehicle control under wave-induced disturbances.
\end{abstract}


\section{Introduction}

\IEEEPARstart{A}{dvanced} control of marine vehicles is sharply becoming an industrial necessity rather than an academic exercise, with the offshore sector seeking higher levels of autonomy with respect to intervention tasks, inspection tasks and similar \cite{Hastie2018}. An industry section where increased autonomy would be highly beneficial is offshore, in particular the marine renewable sector, as the transition to cleaner energy sources begins to accelerate and harsher environments become the subject of exploration for power generation \cite{Zereik2018}. Remotely Operated Vehicles have become a solidified aspect of many subsea procedures, but operation near the free-surface or in shallower waters remains a challenge when the sea state is not calm, owing to the significant influence of surface waves which often limits the deployment of piloted strategies \cite{Khalid2022}. Specifically with respect to marine renewable devices, situation within a turbulent environment is critical to generate sufficient power; thus, a greater level of robustness, reliability and precision is required with regards to vehicle control if autonomous operation is to become a reality \cite{Sivcev2018}.

Attempts to develop automatic disturbance compensation control algorithms have varied, with early implementations investigating vision-based solutions \cite{Leabourne1997, Marks2002}. In calmer environments these strategies can be effective, but there is a lack of robustness due to the dependence on visibility and being feedback based. To tackle this, alternative solutions have suggested inherently robust methods in the form of sliding mode and adaptive controllers \cite{Cui2016, Elmokadem2017}, which account for hydrodynamic loading by considering this as a set of generalised system disturbances. For small-magnitude disturbances this can improve performance sufficiently, but a question remains over stability and performance guarantees when loading increases, for example low-depth operation under large wave heights. 

Handling time-varying and unsteady disturbances remains an open challenge, but a proposition which holds promise lies in exploiting forecasting methods \cite{Fusco2009, Belmont2014}; for wave-induced disturbances which are largely predictable in nature, utilising preview information can assist in disturbance compensation \cite{Fang2022}. Given that wave predictions can be deduced through time-history data, this lends itself to being applicable to a wider range of disturbances than other proposals. Furthermore, the development of state estimation methods \cite{Soylu2016} coupled with a hydrodynamic loading model \cite{WalkerICRA, WalkerRAL} allows the formulation of a feed-forward (FF) control action to reduce state error, in conjunction with an establish feedback controller for set-point regulation. Likewise, the formulation of a single additional control action ensures computation requirements remain low, a major benefit in relation to alternative predictive solutions \cite{Fernandez2017} for these highly dynamic scenarios.

With respect to the above, this work proposes the use of a Feed-Forward (FF) control element which models the wave disturbances as a product of added inertia and hydrodynamic drag, coupling this with a Cascaded Position-Velocity Proportional-Derivative Controller (C-PD) to counteract wave disturbances. The vehicle state is estimated through an Extended Kalman Filter (EKF), which determines the corrective control action in the feedback loop. The controller is simulated under several wave conditions when the disturbance estimation is both well-known and features significant uncertainty, showing that the proposed FF+C-PD controller outperforms a standard C-PD controller in all cases. These findings demonstrate that even coarse predictions of environmental disturbances can prove useful in mitigating state error of subsea vehicles, thus improving the accuracy of dynamic control tasks.

\section{Modelling}

Throughout this work, the vehicle is considered to possess 3DoF and is restricted to a planar case; therefore the presented model considers the surge, heave and pitch motions only. Analogously, the ocean waves are modelled according to 2nd-order planar theory and are assumed to propagate uni-directionally.

\subsection{Vehicle Rigid-Body Dynamics}

\begin{figure}[t!]
    \centering
    \includegraphics[width=0.48\textwidth]{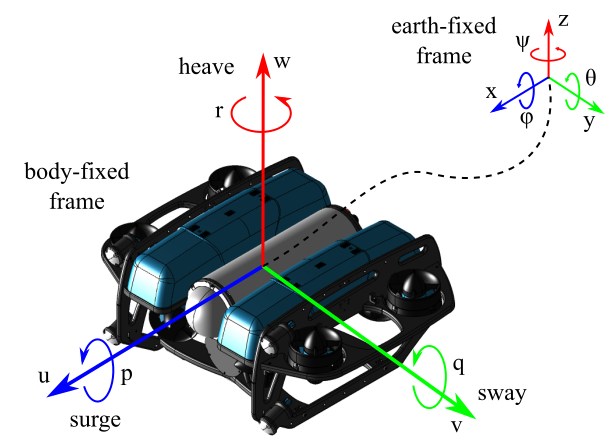}
    \caption{The two frames, earth-fixed and body-fixed, and the relating transformation for the BlueROV2 Heavy configuration.}
    \label{brov2_frames}
\end{figure}

The kinematics of the rigid-body are described by considering two co-ordinate frames; the earth-fixed and body-fixed. As depicted in Fig. \ref{brov2_frames}, these are related by a transformation according to \cite{FossenBook}:
\begin{equation}
    \dot{\boldsymbol{\eta}} = \mathbf{J(\boldsymbol{\eta})}\boldsymbol{\nu}
\end{equation}
\noindent where $\boldsymbol{\eta}$ is a state vector describing the position and orientation of the vehicle, $\boldsymbol{\nu}$ is a state vector of linear and angular velocities and $\mathbf{J}\in \mathbb{R}^{3\times 3}$ is the transformation matrix relating the two frames. 

Given the above kinematic representation, the vehicle dynamics exhibit nonlinear behaviour and are defined according to:
\begin{equation} \label{dynamic_equation}
    \mathbf{M\dot{\boldsymbol{\nu}}} + \mathbf{C(\boldsymbol{\nu})\boldsymbol{\nu}} + \mathbf{D(\boldsymbol{\nu})\boldsymbol{\nu}} + \mathbf{g(\boldsymbol{\eta})} = \boldsymbol{\tau} + \boldsymbol{\tau_{E}}
\end{equation}
\noindent where $\mathbf{M} = \mathbf{M}_{RB} + \mathbf{M}_{A} \in \mathbb{R}^{3\times3}$ is an inertia matrix, $\mathbf{C(\boldsymbol{\nu})}  = \mathbf{C_{RB}(\boldsymbol{\nu})} + \mathbf{C_{A}(\boldsymbol{\nu})} \in \mathbb{R}^{3\times3}$ is a matrix of Coriolis and centripetal terms, $\mathbf{D(\boldsymbol{\nu})} = \mathbf{D}_L(\boldsymbol{\nu}) + \mathbf{D}_Q(\boldsymbol{\nu}) \odot |\boldsymbol{\nu}| \in \mathbb{R}^{3\times3}$ is a hydrodynamic damping matrix and $\mathbf{g(\boldsymbol{\eta})} \in \mathbb{R}^{3}$ is a vector of hydrostatic restoring forces. In the above, subscripts $_{RB}$ and $_A$ relate to contributions from rigid body and added inertial effects, with:
\begin{equation} \label{mass_params}
    \mathbf{M}_{RB} = \begin{bmatrix}
        m & 0 & 0 \\
        0 & m & 0 \\
        0 & 0 & I_{y}
    \end{bmatrix} \quad 
    \mathbf{M}_{A} = \begin{bmatrix}
        X_{\dot{u}} & 0 & X_{\dot{q}} \\
        0 & Z_{\dot{w}} & 0 \\
        M_{\dot{u}} & 0 & M_{\dot{q}}
    \end{bmatrix}
\end{equation}
where $m$ is the vehicle dry mass, $I_{y}$ is a rotational inertia and $ X_{\dot{q}}$, $Z_{\dot{w}}$, and $X_{\dot{u}} = M_{\dot{u}}$ are added mass coefficients. The Coriolis and centripetal terms are derived in accordance with Eq. \ref{mass_params} \cite{FossenBook} and the damping matrix is specified as:
\begin{equation} \label{damping_params}
    \mathbf{D}_{L} = \begin{bmatrix}
        X_u & 0 & 0 \\
        0 & Z_w & 0 \\
        0 & 0 & M_q
    \end{bmatrix} \quad 
    \mathbf{D}_{Q} = \begin{bmatrix}
        X_{u|u|} & 0 & 0 \\
        0 & Z_{w|w|} & 0 \\
        0 & 0 & M_{q|q|}
    \end{bmatrix}
\end{equation}

where $X_{u}$, $Z_{w}$ and $M_q$ are linear damping coefficients whilst $X_{u|u|}$, $Z_{w|w|}$ and $M_{q|q|}$ are quadratic damping coefficients. All hydrodynamic parameters are defined in Table \ref{BlueROV2_Parameters}. 

Finally, $\boldsymbol{\tau}\in \mathbb{R}^{3}$ is a vector of control forces and moments whilst environmental disturbances are lumped within the vector $\boldsymbol{\tau_{E}}\in \mathbb{R}^{3}$. As this work is concerned with mitigating the effects of surface waves on vehicle behaviour, disturbances arising from ocean currents are assumed negligible and $\boldsymbol{\tau_{E}}$ is formulated to purely describe wave-induced loading, according the model presented in the following section.

\subsection{Wave-Induced Disturbances}

\begin{figure*}[t!]
    \centering
    \includegraphics[width=.95\textwidth]{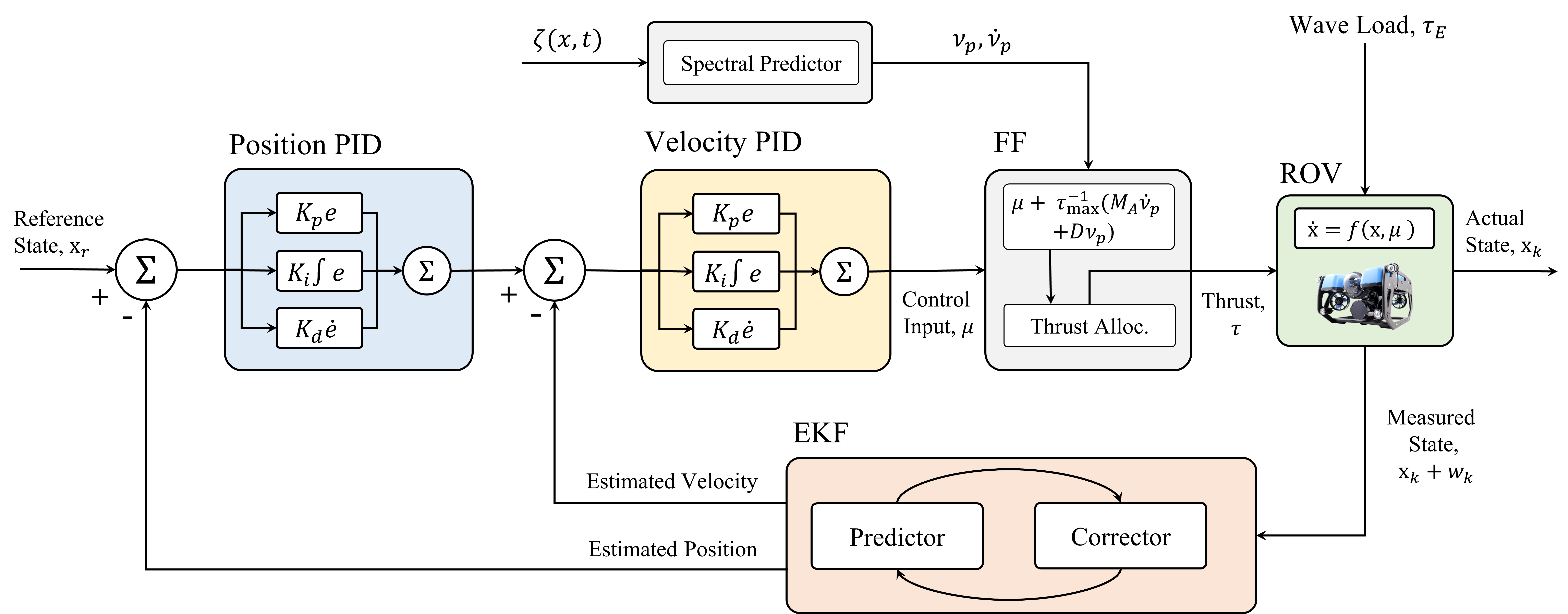}
    \caption{Block diagram of the proposed feed-forward disturbance mitigation technique, with the grey blocks representing the generation of the additional compensating control actions.}
    \label{block_diagram}
\end{figure*}

To form a temporal history of the wave elevation, a 2nd-order model was adopted utilising the principle of superposition. The sea state is therefore described by a spectrum of $N$ monochromatic components, each with a unique wave amplitude $A$, wave period, $T$, and phase offset $\epsilon$. It follows that at a point $x$ and at time $t$ the wave elevation is described by \cite{FossenBook}:
\begin{align}
\zeta(x, t) = & \sum_{i=0}^{N} A_{i}cos(\kappa_{i}x - \omega_{i} t + \epsilon_{i}) \nonumber \\
  & + \sum_{i=1}^{N} \frac{1}{2}\kappa_{i}A_{i}^{2} cos2(\kappa_{i}x - \omega_{i} t + \epsilon_{i})
\label{WaveEquation}
\end{align}
where $\kappa$ and $\omega$ are the wave number and circular frequency respectively. These additional parameters are deduced by solving the dispersion relation:
\begin{equation}
    \omega_{i} = \sqrt{g\kappa_{i} \tanh{\kappa_{i}d}}
\end{equation}
where $g$ and $d$ are the gravitational constant and approximate seabed depth. In Eq. \ref{WaveEquation}, the wave amplitude is related to the spectral density function $S(\omega)$ by $A^{2}=2S(\omega)\Delta \omega$ where $\Delta \omega$ is the difference between successive frequencies. Several models exist to describe the spectral density function \cite{FossenBook}; in this work the JONSWAP spectrum is adopted owing to the relation with the North Sea, an area of interest for this application. The spectral density function is therefore:
\begin{equation}
    S(\omega) = \frac{\alpha g^{2}}{\omega^{5}} \exp \left[ -\frac{5}{4} \left( \frac{\omega_{p}}{\omega} \right)^{4} \right] \gamma^{\Gamma}
\end{equation}
where $\omega_{p}$ is the spectral peak frequency, $\gamma^{\Gamma}$ is a peak enhancement factor and
\begin{gather}
    \alpha =  0.0081  \nonumber \\
    \sigma = \begin{cases}
      0.07, & \text{if}\ \omega \leq \omega_{p} \\
      0.09, & \text{if}\ \omega \geq \omega_{p}
    \end{cases} \nonumber\\
    \Gamma = \exp \left[ \frac{(\omega - \omega_{p})^{2}}{2\omega_{p}^{2}\sigma^{2}}\right] \nonumber
\end{gather}

Spectral information can also be exploited to deduce the fluid particle motions at a point beneath the surface $z$, which for 2nd-order theory produces \cite{McCormickBook}:
\begin{align}
	\label{ParticleVelocityX}
	u_{p}(x, z, t) = & \sum_{i=1}^{N} \frac{g H_{i}}{2c}\frac{\cosh \kappa_{i}(z+d)}{\cosh \kappa_{i}d} \cos(\kappa_{i}x - \omega_{i} t + \epsilon_{i}) + \nonumber\\
        & \frac{3}{16}c\kappa_{i}^2H_{i}^2 \frac{\cosh[2\kappa_{i}(z+d)]}{\sinh^{4} \kappa_{i}d} \cos[2(\kappa_{i}x - \omega_{i} t + \epsilon_{i})]
\end{align}	
\begin{align}
	\label{ParticleVelocityZ}
	w_{p}(x, z, t) = & \sum_{i=1}^{N} \frac{g H_{i}}{2c}\frac{\sinh \kappa_{i}(z+d)}{\cosh \kappa_{i}d} \sin(\kappa_{i}x - \omega_{i} t + \epsilon_{i}) + \nonumber\\
        & \frac{3}{16}c\kappa_{i}^2H_{i}^2 \frac{\sinh[2\kappa_{i}(z+d)]}{\sinh^{4} \kappa_{i}d} \sin[2(\kappa_{i}x - \omega_{i} t + \epsilon_{i})]
\end{align}
where $c$ is celerity, defined according to seabed depth to wavelength ratio \cite{ReeveBook}. This facilitates deduction of the wave-induced hydrodynamic loads ($X_{E}$, $Z_{E}$ and $M_{E}$ for the surge, heave and pitch respectively) acting on the body; here, a low-order model is employed which has been experimentally validated in previous work \cite{WalkerRAL, WalkerICRA, Gabl2020, Gabl2021}:
\begin{equation} \label{wave_forces}
    \boldsymbol{\tau}_{E} = \begin{bmatrix}
        X_{E} \\ Z_{E} \\ M_{E}
    \end{bmatrix} = 
    \begin{bmatrix}
        X_{\dot{u}}\dot{\nu}_{p,x} + \lbrace X_{u} + X_{u|u|}|\nu_{p,x}| \rbrace \nu_{p,x} \\ Z_{\dot{w}}\dot{\nu}_{p,z} + \lbrace Z_{w} + Z_{w|w|}|\nu_{p,z}| \rbrace \nu_{p,z} \\ 
        \int_{-L/2}^{L/2} Z_{E}(x',z',t)x' dx
    \end{bmatrix}
\end{equation}
where $\boldsymbol{\nu}_{p} = \left[ \nu_{p,x}, \nu_{p,z} \right]^{T} = \mathbf{R}_{y}(\theta)\left[ u_{p}, w_{p} \right]^{T}$ ($\mathbf{R}_{y}(\theta)$ is a rotation matrix). Also, $L$ is the vehicle body length and $(x',z')$ refers to points along the vehicle axial length within the local frame. 

\section{Control Methodology}

Obtaining spectral knowledge of the immediate ocean environment around the vehicle is key in deducing the magnitude of disturbances when applying the model described by Eq. \ref{wave_forces}. Several methods have been proposed for predicting impending waves and wave loads, including but not limited to \emph{in-situ} sensor fusion \cite{Selvakumar2016}, auto-regressive models \cite{Fusco2009} and deterministic methods \cite{Belmont2014}, the latter in particular focusing on utilising spectral information. These have been applied successfully in the context of wave energy converters, therefore an analogous deployment with respect to an underwater vehicle holds adjacent potential. It is therefore postulated here that exploiting knowledge of wave disturbances to formulate a feed-forward control action can assist in reducing state perturbations, improving station keeping performance and widening the range of deployable conditions. The control law is therefore formulated as a Cascaded Proportional-Derivative (C-PD) with feed-forward (FF) action: 
\begin{align}
    \boldsymbol{\tau} = & \underbrace{\boldsymbol{\tau}_{max} \left[ \mathbf{K}_{p,v} \lbrace \boldsymbol{\nu} - \left( \mathbf{K}_{p}\mathbf{e} + \mathbf{K}_{d}\mathbf{\dot{e}} \right) \rbrace \right] }_{C-PD}  \\
    & + \underbrace{ \lbrace \mathbf{M}_{A}\boldsymbol{\dot{\nu}}_{p} + \mathbf{D}(\boldsymbol{\nu_{p}})(-\boldsymbol{\nu}_{p})\rbrace}_{FF} \nonumber
\end{align}
where $\mathbf{K}_{p}$ and $\mathbf{K}_{d}$ are the position PD gains, $\mathbf{K}_{p,v}$ is the velocity P-gain and $\mathbf{e}$ is the state error. Also, $\boldsymbol{\tau}_{max} \in \mathbb{R}^{3}$ is a vector describing the maximum torque available in each DoF. As the heave and pitch states are controlled via the horizontal thrusters, a thrust allocation algorithm is embedded within the control architecture to generate the appropriate control inputs to deliver the required forces and moments. Inclusion of the motor time-delay as a first-order response yields:%
\begin{equation} \label{inverse_control}
    \boldsymbol{\mu} = \left[ \left( 1-e^{-\Delta t / t_{m}} \right)\mathbf{K}_{\boldsymbol{\tau}}^{-1} \right] \boldsymbol{B}_{\boldsymbol{\mu}}^{\dagger}\boldsymbol{\tau}
\end{equation}
where $\boldsymbol{B}_{\boldsymbol{\mu}}^{\dagger}$ is the Moore-Penrose pseudo-inverse of the thrust allocation matrix $\boldsymbol{B}_{\boldsymbol{\mu}}$, $\mathbf{K}_{\boldsymbol{\tau}}$ is a force co-efficient matrix and $\Delta t_{m}$ is the motor-time constant. Eq. \ref{inverse_control} produces a solution $\boldsymbol{\mu} \in \mathbb{R}^{8}$ to be allocated to each thruster.

\begin{figure}[t!]
    \centering
    \begin{subfigure}[t]{0.244\textwidth}
        \centering
        \includegraphics[width=1\textwidth]{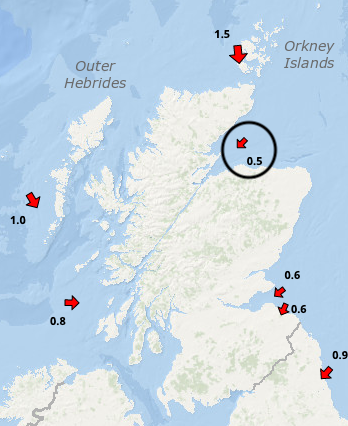}
        \caption{}
    \end{subfigure}%
    ~ 
    \begin{subfigure}[t]{0.232\textwidth}
        \centering
        \includegraphics[width=1\textwidth]{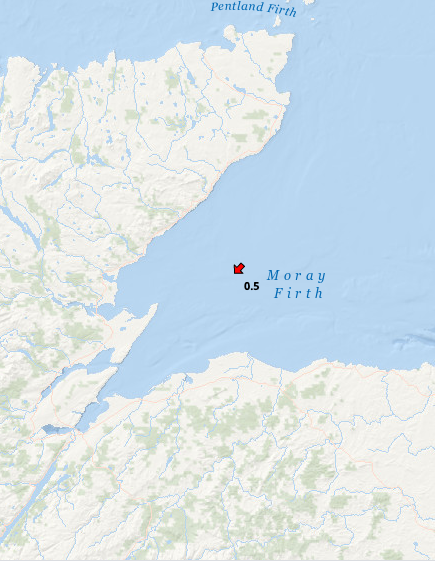}
        \caption{}
    \end{subfigure}
    \caption{Map showing the (a) various locations of buoys around Scotland, circling the buoy selected for this analysis and (b) an enlarged map of the Moray Firth buoy location.}
    \label{buoy_location}
\end{figure}

\subsection{State Estimation}
Deploying automatic control in practice requires knowledge of the vehicles location to be known or at minimum a reasonable estimation to be inferred; to add an additional layer of realism to the simulations, we employ an EKF as a state estimator to track the vehicle state. The EKF algorithm has two update phases; the \emph{predictor} phase and the \emph{corrector} phase. The predictor phase considers the initial estimates and projects the error covariance and state, $\mathbf{P}$ and $\mathbf{\hat{x}}$, ahead in time such that:
\begin{equation} \label{error_covariance_predict}
    \mathbf{P}_{k}^{-} = \mathbf{A}_{k}\mathbf{P}_{k-1}\mathbf{A}_{k}^{T} + \mathbf{Q}_{k-1}
\end{equation}
\begin{equation}
    \mathbf{\hat{x}}_{k}^{-} = F(\hat{\mathbf{x}}_{k-1},\boldsymbol{\mu_{k-1}},\mathbf{w})
\end{equation}
where $\mathbf{A} = \frac{dF}{dx} \rvert_{\mathbf{x} = \mathbf{\hat{x}}} $ is the linearised state transition matrix, $\mathbf{Q}$ is the process error covariance and $\mathbf{w}$ represents the process noise. Here, $F$ represents a nonlinear function for the vehicle dynamics. From this, the corrector phase proceeds to compute the Kalman gain:
\begin{equation}
    \mathbf{K}_{k} = \mathbf{P}_{k}^{-}\mathbf{H}_{k}^{T}(\mathbf{H}_{k}\mathbf{P}_{k}^{-}\mathbf{H}_{k}^{T} + \mathbf{R}_{k})^{-1}
\end{equation}
where $\mathbf{H}_{k}$ is a positional measurement and $\mathbf{R}_{k}$ is the measurement error covariance, before updating the estimate with a measurement $y_{k}$
\begin{equation}
    \mathbf{\hat{x}}_{k} = \mathbf{\hat{x}}_{k}^{-} + \mathbf{K}_{k}(y_{k} -\mathbf{H}_{k}\mathbf{\hat{x}}_{k}^{-})
\end{equation}
Similarly, the error covariance is also corrected in this stage before looping back to Eq. \ref{error_covariance_predict} and repeating for every time-step $k$, where $\mathbf{I}$ is an identity matrix:
\begin{equation}
    \mathbf{P}_{k} = (\mathbf{I}-\mathbf{K}_{k}\mathbf{H}_{k})\mathbf{P}_{k}^{-}
\end{equation}

\begin{table}[t!] 
\renewcommand{\arraystretch}{1.3}
\caption{Case assignments and parameters for the analysed wave spectra.}.
\label{spectral_parameters}
\centering
\begin{tabular}{|c|c|c|} 
\hline
Case Reference  & Peak Period (s) & Significant Wave Height (m) \\
\hline
W1 & 7.1 & 2.78\\
W2 & 9.5 & 3.47 \\
W3 & 11.1 & 3.24 \\
\hline
\end{tabular}
\end{table}

\begin{table}[t!]
\renewcommand{\arraystretch}{1.3}
\caption{BlueROV2 Heavy dimensions and hydrodynamic parameters; data based on \cite{Benzon2021, brov2, Wu2018}.}
\label{BlueROV2_Parameters}
\centering
\begin{tabular}{|c|c|c|} 
\hline
Parameter & Nomenclature & Value \\
\hline
Weight & $W$ & 112.8 N\\
Buoyancy & $B$ & 114.8 N\\
Rotational Inertia, $y$ & $I_{y}$ & 0.253 kgm$^{2}$ \\
Added Inertia Coeff. & $X_{\dot{u}}$, $Z_{\dot{w}}$ & 6.36, 18.68 kg\\
" & $M_{\dot{q}}$ & 0.135 kgm$^{2}$\\
" & $X_{\dot{q}}$, $M_{\dot{u}}$ & 0.67 kgm \\
Linear Drag Coeff. & $X_{u}$, $Z_{w}$ & 13.7, 33 kg/s\\
" & $M_{q}$ & 0.80 kgm$^{2}$/s\\
Quadratic Drag Coeff. & $X_{u|u|}$, $Z_{w|w|}$ & 141, 190  Ns$^{2}$/m$^{2}$\\
" & $M_{q|q|}$ & 0.47 Nms$^{2}$ \\
Centre of Buoyancy & $r_{B}$ & [0, 0, 0.028]m \\
Maximum Thrust & $T_{max}$ & 35 N \\
Thruster Offset & $\alpha$ & 45$^{o}$ \\
\hline
\end{tabular}
\end{table}

\section{Scenario Configuration}

Given the intended application of the proposed disturbance mitigation method is for improved performance during inspection and maintenance of devices/structures in wave-dominated environments, the vehicle was simulated under three different sea states to analyse performance relative to varying wave parameters. Spectral data was sourced from the online repository of the Centre for Environmental Fishes and Aquaculture Science (Cefas) \cite{cefas}, collected by a wave buoy situated off the coast of Inverness, Scotland in the Moray Firth. The location of the buoy is shown in Fig. \ref{buoy_location} where $d=54m$; offshore wind farms are typically located within areas of this depth \cite{Bailey2014, Oh2018}, thus emulating the conditions of a typical inspection or maintenance task. The selected wave fields and assigned case references are given in Table \ref{spectral_parameters} with all vehicle parameters given in Table \ref{BlueROV2_Parameters}.

The controller was tasked with performing station-keeping at a depth of $z=5$m and for a temporal segment of $600$s with a resolution of $\Delta t = 0.05s$, exposing the vehicle to significant magnitude wave disturbances for a prolonged period of time. It should be noted here that station-keeping refers to both positional and attitude regulation, during which the controller attempts to maintain a reference set-point $\mathbf{x}_r = [x_{r}, z_{r}, \theta_{r}]^{T}$.  Throughout all analysis, sensor noise is considered which is mitigated by the inclusion of an EKF to estimate the vehicle state. Comparisons are drawn between the FF controller and a standard C-PD controller as a baseline reference which also exploits the EKF to monitor state error. 

\begin{figure}[t!]
    \centering
    \includegraphics[width=0.48\textwidth]{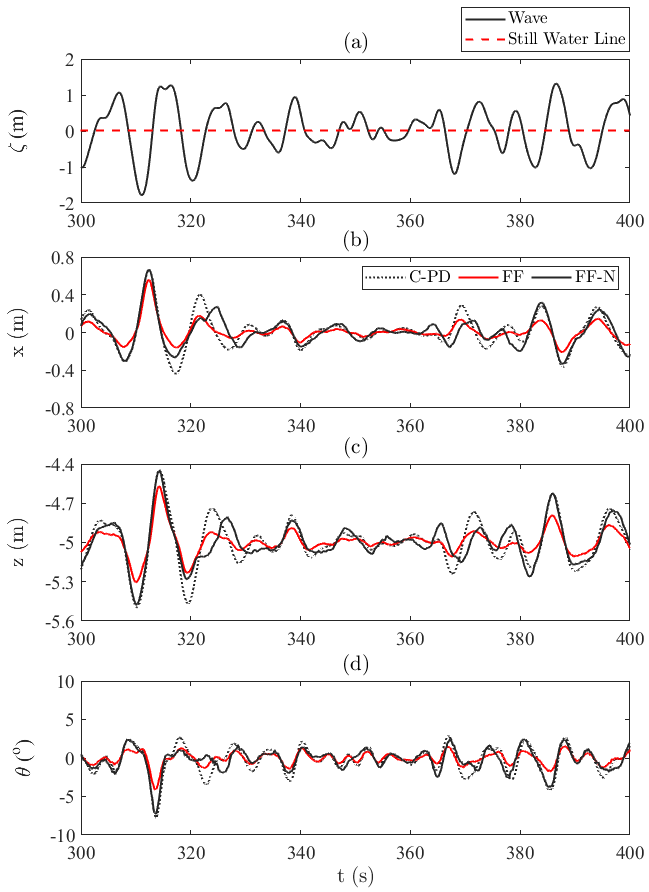}
    \caption{Temporal segment of the (a) wave in case W3. showing the (b) surge position, (c) heave position and (d) pitch attitude.}
    \label{time_history}
\end{figure}

\section{Simulation Results}

Performance of the proposed control scheme was analysed by considering the Root-Mean-Square-Error (RMSE) and absolute maximum error witnessed across the 600s simulation; the power consumed during the station keeping mission was also recorded and analysed.

\subsection{Station Keeping Performance}

A temporal segment for case W3 is displayed in Fig. \ref{time_history} which evidently shows the FF compensation having a positive impact on the station keeping error, relative to the C-PD controller. The preview knowledge of the wave is seen to reduce RMSE by up to $58.6\%$, with this specific value relating to the pitch motion for case W1. For the surge and heave, the maximum reductions were also related to case W1 which returned a $47.5\%$ and $48.2\%$ improvement respectively. This is an indicator that for lower peak period spectra the inclusion of FF compensation is more critical, however this could be owed to the causal effect of larger pitch perturbations in contrast to cases W1 and W2. 

In terms of the maximum displacement recorded during the simulations, the greatest improvement was also related to the pitch, reducing attitude error by $50.1\%$ on average. This was much lower for the other DoF, with the surge and heave only showing $22\%$ and $11.8\%$ reductions; it is suspected that this is due to a brief section of the simulated waves subjecting the vehicle to sharp, high magnitude disturbances which the control was unable to correct for. Behaviour similar to this can be seen in Fig. \ref{time_history} at $\approx 315$s, where the traces for the control scheme undergo similar magnitude displacements for the largest wave height in the segment.  Across all cases there was a mean reduction in RMSE of $\approx 48\%$ and $\approx 28\%$ in maximum error; all absolute values are shown in Fig. \ref{errors}.

\subsection{Sensitivity to Noise}

\begin{figure}[t!]
    \centering
    \includegraphics[width=0.48\textwidth]{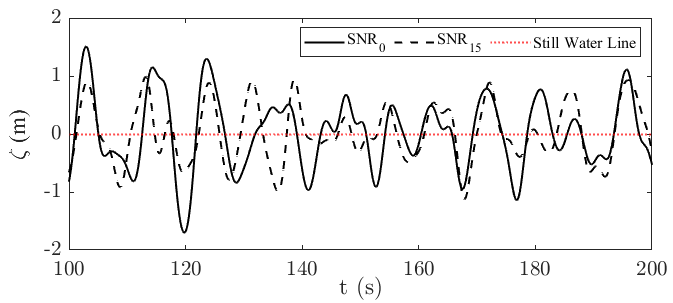}
    \caption{Temporal segment for case W3 during noise analysis; here, SNR$_{0}$ is the encountered wave by the vehicle where as SNR$_{15}$ is the expected wave by the controller used to generate compensating control actions. }
    \label{snr_waves}
\end{figure}

\begin{figure}[t!]
    \centering
    \includegraphics[width=0.48\textwidth]{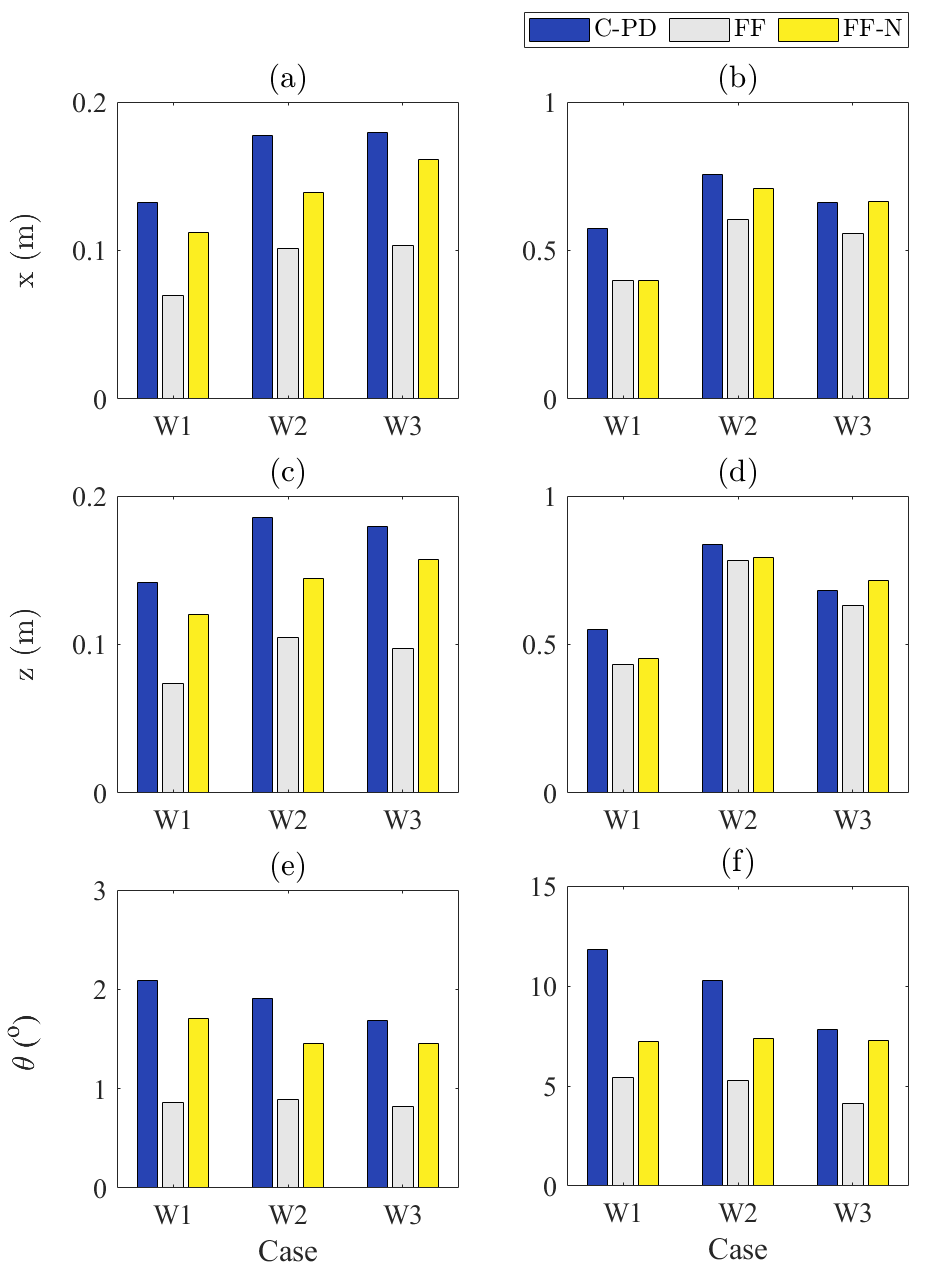}
    \caption{State error for each control method during the station keeping simulation, showing the (a)(c)(e) RMSE and (b)(d)(f) absolute maximum error recorded.}
    \label{errors}
\end{figure}

Each case was analysed when both the disturbance feed-forward term is deemed to be accurate (denoted SNR$_{0}$) and feature imprecisions (denoted SNR$_{15}$); the latter attempts to provide insight into controller performance when the vehicle encounters disturbances that differ from those anticipated by the FF controller. To achieve this, spectral noise with a SNR of $15$ was injected directly to the spectral component amplitude and phase offset to alter the wave (and thus FF compensation calculations) significantly and randomly as shown in Fig. \ref{snr_waves}. These results are also compiled within Fig. \ref{errors}.

When the disturbance is considered to be inaccurate, there is still improvement in station keeping accuracy across all cases with respect to RMSE and the majority of cases with respect to maximum error; only case W3 returns higher maximum error in the surge and heave and in these specific instances the difference is marginal with increases of $<1\%$ and $5\%$ respectively. The average reduction in RMSE of $17\%$ supports the claim that these instances are at isolated periods in time and not a regular occurrence; similar to above, the pitch experienced the highest improvement in error of $14.2\%$. Overall these results clearly show that even utilising a spectral estimation that features inaccuracies can offer a noticeable improvement in state regulation, implying that well established methods can be applied with confidence at this end of the control pipeline.

\subsection{Station Keeping Efficiency}

\begin{figure}[t!]
    \centering
    \includegraphics[width=0.48\textwidth]{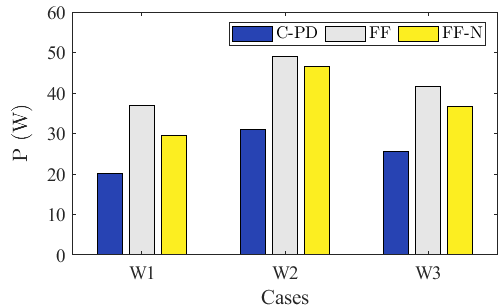}
    \caption{Power consumed during the station keeping mission for each case and control method.}
    \label{power_bar}
\end{figure}

Given the improvement in performance and additional control action generated by the FF component, it was anticipated additional power would be required. To confirm this, the power consumed during the mission was modelled according to manufacturer data such that \cite{T200}:
\begin{equation}
    \mathbf{P} = 0.0011\boldsymbol{\tau}^{3} + 0.02078\boldsymbol{\tau}^{2} + 0.297\boldsymbol{\tau}
\end{equation}
where $\mathbf{P} \in \mathbb{R}^3$ is the power consumed by the control actions in each DoF. This was modelled according to a nominal operating volage of 16V and the results are displayed in Fig. \ref{power_bar}.

The power does increase when employing the FF scheme, but interestingly less power was consumed when considering noisy disturbance estimations. It is likely that this is attributed to the FF controller anticipating lower magnitude disturbances for the majority of the simulated mission, which would explain the increase in error that is shown in Fig. \ref{errors}. Similarly, the segment displayed in Fig. \ref{snr_waves} demonstrates this behaviour at various instances, for example at 120s where the disturbance is significantly underestimated. This supports the connected data in relation to the positional and attitude RMSE/maximum error. For large magnitude waves it can be argued that this increased power expenditure is utilised well to reduce station keeping error significantly, thus in situations where this is paramount it becomes less of a burdening factor. 

\section{Conclusions}

This paper has proposed active wave-induced disturbance rejection for underwater vehicles based on the inclusion of a feed-forward control action. An extensive simulation study demonstrated the capability of the proposed control to significantly reduce state regulation error in both position and attitude, highlighting performance improvements of up to $56.8\%$ over three sea states with varying parameters. Similarly, the maximum error (which is arguably the critical factor in these scenarios) was also reduced substantially, with more prominent reductions witnessed for the vehicle pitch. Given the availability of wave prediction tools, these results provide evidence of the real potential related to the incorporation of modelled wave-induced loads directly within the control to improve performance. Considering this, an interesting avenue for exploration would be to develop a preview of wave-induced disturbances along a future time-horizon; this would facilitate the development of different forms of model predictive control, potentially improving performance even further by evaluating an optimal control sequence, rather than a one step control action. 




\end{document}